\documentclass[]{llncs}

\usepackage[T1]{fontenc}

\usepackage{graphicx}
\usepackage{amsmath,amssymb,amsfonts}
\usepackage{algorithmic}
\usepackage{textcomp}
\usepackage{xcolor}
\usepackage{tikz}
\usepackage{placeins}
\usepackage{booktabs}
\usepackage{tabularx}
\usepackage{float}
\usetikzlibrary{positioning}

\begin{document}

\title{Modeling Decisions in Blockchain Analytics: A Leakage-Aware Evaluation of Tree-Based vs. Sequential Models}
\titlerunning{Modeling Decisions in Blockchain Analytics}

\author{Michał Bartnicki\inst{1}\orcidID{0009-0004-5856-4850} \and
Jarosław A. Chudziak\inst{1}\orcidID{0000-0003-4534-8652}}
\institute{Faculty of Electronics and Information Technology \\ Warsaw University of Technology, Warsaw, Poland \\
\email{\{michal.bartnicki.stud, jaroslaw.chudziak\}@pw.edu.pl}}
\authorrunning{M. Bartnicki and J. A. Chudziak} 

\maketitle

\begin{abstract}
Sybil bots are Ethereum actors that imitate legitimate users to extract airdrop rewards or influence governance. Recent Sybil detection methods increasingly use deep learning and treat blockchain activity as a quasi-linguistic sequence.
However, complex sequence models are computationally expensive for real-time monitoring, and their reported performance may be inflated by label leakage from high-signal smart contracts.
We ask whether and how organic users, Sybil bots, and MEV bots differ in the structural complexity of their transaction histories; whether sequential models outperform tree-based tabular models once leakage is reduced; whether transaction order or timing provides the stronger behavioral signal; and whether the resulting models are practical for low-latency deployment. Our approach to leakage-aware Sybil bot detection consists of a Blind-Spot protocol and a Transaction Grammar representation of wallet behavior. The former eliminates shortcuts associated with high-signal contracts, whereas the latter models wallets using rhythm, EVM execution structure, and intent. We evaluate this approach on Ethereum actor classification by comparing Transformer and BiLSTM sequence models against XGBoost and SVM baselines. We contribute a framework for leakage-aware Ethereum actor classification and a Transaction Grammar representation of wallet behavior. Our results demonstrate that, under leakage-aware evaluation, XGBoost outperforms Transformer-based sequence models while providing lower latency and estimated energy use.

\end{abstract}

\section{Introduction}

Blockchain systems increasingly rely on automated actors to provide liquidity and arbitrage. While Maximal Extractable Value (MEV) bots, known as arbitrage bots, are generally an expected component of these ecosystems \cite{Materwala_2025}, an emerging concern is the growing number of automated actors masquerading as legitimate users – Sybil bots. This challenges fairness and incentive design, as Sybil farmers undermine airdrops and governance mechanisms intended for legitimate participants \cite{messias2025airdropsgivingmoneyaway}. With the saturation of legitimate MEV activity, distinguishing between Sybil behavior and essential MEV infrastructure has become critical.

Recent trends in blockchain behavioral modeling emphasize complex deep learning, particularly Natural Language Processing (NLP)-inspired sequence models \cite{SUN2025103074}. These approaches operate on the assumption that transaction history resembles a language exhibiting meaningful long-range dependencies \cite{liu2024roletransformermodelsadvancing}. However, this assumption lacks rigorous evaluation. Unlike natural language, transaction history is sparse, highly structured, and often dominated by aggregate statistics rather than compositional semantics. This aligns with recent findings suggesting that in highly volatile financial domains, a strategic subset of features, or a partial-multivariate approach, often offers superior predictive power over noise-prone full-multivariate models \cite{tokajuk2025partialmultivariatetransformertool}. Notably, many state-of-the-art results rely on superficial interactions with specific contracts, introducing label leakage that inflates reported performance.

In this work, we revisit blockchain actor classification under a leakage-aware setting. To this end, we introduce the \emph{Blind-Spot} protocol, which mitigates the influence of high-signal smart contracts. To better capture intrinsic patterns, we propose a hierarchical \emph{transaction grammar} that parses transaction sequences using the internal \emph{Ethereum Virtual Machine (EVM) execution trace} of every call. We leverage this framework to compare sequential and tabular inductive biases under identical information conditions. We investigate whether organic and automated actors diverge grammatically and whether sequential models retain their advantage once shortcut features are removed.

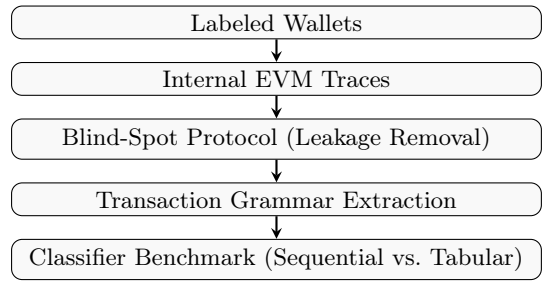
\begin{figure}[htbp]
\centering
\begin{tikzpicture}[
    node distance=0.3cm,
    every node/.style={draw, rectangle, rounded corners, align=center, font=\footnotesize, fill=gray!5, minimum width=7cm},
    arrow/.style={->, >=stealth, thick}
]
\node (labels) {Labeled Wallets};
\node (tx) [below=of labels] {Internal EVM Traces};
\node (leak) [below=of tx] {Blind-Spot Protocol (Leakage Removal)};
\node (repr) [below=of leak] {Transaction Grammar Extraction};
\node (model) [below=of repr] {Classifier Benchmark (Sequential vs. Tabular)};

\draw[arrow] (labels) -- (tx);
\draw[arrow] (tx) -- (leak);
\draw[arrow] (leak) -- (repr);
\draw[arrow] (repr) -- (model);
\end{tikzpicture}
\caption{End-to-end workflow for leakage-aware Ethereum actor classification.}
\label{fig:workflow}
\end{figure}

We find that while organic users exhibit higher structural entropy and complexity than automated actors, attention-based models exploit this structure only weakly. In contrast, tree-based models outperform deep learning architectures at a fraction of the computational cost, highlighting a critical efficiency-performance trade-off for sustainable, real-time blockchain analytics. This finding aligns with broader literature demonstrating that tree ensembles often retain state-of-the-art performance on tabular domains where deep learning inductive biases are ill-suited \cite{Borisov_2024}. We illustrate our end-to-end workflow in Fig. \ref{fig:workflow}.

\section{Related Work}
\label{sec:related_work}

\paragraph{Blockchain actor and security classification}
Researchers have explored neural networks for blockchain analytics, including sequence encoder models for transaction histories \cite{liu2025detectingsybiladdressesblockchain} and graph neural networks for interaction graph structures \cite{Niedermayer_2024}, while traditional gradient boosting decision trees remain effective baselines due to their strong performance on structured features and moderate-sized datasets \cite{palaiokrassas2024machinelearningblockchaindata}.
However, reported performance is often inflated because models exploit class-defining proxies rather than generalizable patterns \cite{grinsztajn2022treebasedmodelsoutperformdeep}.
 
\paragraph{Deep learning vs. trees on tabular data}
Apart from blockchain, existing research confirms the lack of dominance of deep learning approaches over tree-based models in tabular data domains, where gradient boosting decision trees have been found to obtain higher accuracy with less tuning complexity in large-scale benchmarks \cite{shwartzziv2021tabulardatadeeplearning}. To address these differences, several transformer-based models have been proposed for tabular learning, such as FT-Transformer \cite{gorishniy2023revisitingdeeplearningmodels} and SAINT \cite{somepalli2021saintimprovedneuralnetworks}, where self-attention is utilized for tabular data learning. While these models have been found to be effective, research confirms limited advantages for medium-scale data sets, where the presence of uninformative features is critical for model performance. While pretraining-based models like TabPFN \cite{hollmann2023tabpfntransformersolvessmall} excel on small datasets, their complexity and scalability issues make comparing sequential and tabular inductive biases critical for Ethereum modeling. Furthermore, while Transformer models show promise for financial time series, their performance is heavily contingent on the choice of training objectives and loss functions, as demonstrated in stock ranking tasks \cite{Kwiatkowski_2025}. Efficient long-sequence variants such as Informer address the quadratic attention bottleneck through sparse attention, and have also been applied in financial option pricing, which further motivates verifying whether long-range sequence modeling is actually beneficial in blockchain analytics \cite{zhou2021informer,banka2025applying}.
\paragraph{Leakage-aware evaluation}
Literature emphasizes the need for a robust evaluation and leakage mitigation process, where even benign feature selection can reveal target labels \cite{Geirhos_2020}. We define label leakage as the presence of domain-specific 'shortcuts' (e.g., protocol and counterparty addresses) that trivially expose an actor's class. Consequently, explicit controls are necessary to accurately measure generalization behaviors \cite{FAN2021102587}. As a result of these considerations, we propose a leakage-oriented benchmark for Ethereum actor classification that directly compares sequential and tabular inductive bias under a common evaluation methodology, as well as providing efficiency metrics of interest for real-time deployment.

\section{Proposed Approach}
\label{sec:methodology}

The central methodological question is how the same leakage-aware transaction history should be consumed by competing inductive biases. Accordingly, this section first defines a common Transaction Grammar representation, derived from EVM traces and expressed through rhythm, structure, and intent tokens. Sequential architectures consume the grammar as an ordered transaction sequence, whereas tree-based and linear baselines consume aggregate statistics derived from the same grammar. This makes the title comparison explicit: we test whether preserving order adds predictive value once obvious leakage is removed.

\subsection{Transaction Grammar as a Sequence Representation}
\label{sec:sequence_rep}

We represent each wallet as a sequence of transactions $S_W = \{x_1, \dots, x_L\}$. Each transaction $x_t$ is encoded as a triplet $x_t = (\tau_t, \gamma_t, \nu_t)$ capturing \emph{Rhythm}, \emph{Structure}, and \emph{Intent}.

\textbf{Rhythm ($\tau_t$):} We compute the inter-arrival time $\Delta t$ since $x_{t-1}$ and discretize it into 16 bins, ranging from same-block activity ($\Delta t = 0$) to long-term dormancy ($\Delta t > 30$ days), with a special \texttt{START} token for sequence start.

\textbf{Structure ($\gamma_t$):} We describe each transaction using nine EVM trace properties (e.g., depth and error indicators). Each property is discretized and concatenated into a single composite token (e.g., \texttt{Cnt:2\_Dp:4\_Err:0...}), representing the execution-trace shape.

\textbf{Intent ($\nu_t$):} We encode semantic intent using the 4-byte function selector. To control vocabulary size and reduce class imbalance, we merge the top-$k$ selectors from each class into a global vocabulary of size $|V|=100$, mapping rare selectors to \texttt{[UNK]}.

\subsection{Sequential vs. Tree-Based Model Families}
\label{sec:model_arch}

The Transaction Grammar is shared across both model families. For sequential models, the grammar preserves transaction order and allows attention or recurrence to exploit temporal dependencies. Tree-based and linear baselines aggregate the same grammar into distributional features. We also compare token-level and transaction-level sequence architectures.

\textbf{Flattened Transformer (Token-Level).} We utilize a standard Transformer \cite{vaswani2023attentionneed} encoder on the flattened list of all tokens. The wallet transaction history is transformed from a tensor with $L$ transactions and $K$ tokens in each transaction into a single list with $N=L \times K$ tokens. Each token is encoded and added with learnable positional embeddings. We use 4 layers of the Transformer encoder with dimensions $d_{model}=256$ and $n_{head}=4$ with causal masking disabled. We also add a \texttt{[CLS]} token to aggregate global context for the final linear classifier. This approach enables the attention mechanism to attend to any two tokens, regardless of the transaction they belong to. We also evaluate a Hierarchical Transformer, which aggregates tokens into transaction vectors before sequence encoding.

\textbf{Hierarchical BiLSTM (Transaction-Level).}
For the Recurrent Baseline, a hierarchical approach is adopted to overcome the vanishing gradient problem associated with the processing of long token streams. First, the embeddings are summed within each transaction to form a dense feature vector $h_t$. The set of transaction vectors is then input into a Bidirectional Long Short-Term Memory (BiLSTM) \cite{Longshort} architecture (2-layers, 256 hidden units). The hidden states across all time steps are aggregated via max-pooling and projected to class probability outputs. Unlike the Transformer architecture, sequential dependencies are explicitly modeled at the transactional cadence.

\textbf{Tabular Baselines (Feature Engineering).} To validate the necessity of sequential dependencies in the model, a Gradient Boosted Decision Tree (XGBoost) \cite{Chen_2016} is trained on global aggregate statistics. The feature set is a ``Bag-of-Grammar`` consisting of:
\begin{itemize}
    \item Token Counts: Counting the frequency of all unique Rhythm, Structure, and Intent tokens. This approximates the distribution of time and call types.
    \item Complexity Metrics: Shannon Entropy and Lempel-Ziv (LZ) compression ratios are computed on the token streams to quantify the diversity in behavior.
\end{itemize}
A Linear Support Vector Machine (SVM) is also used as a feature engineering baseline on Term Frequency-Inverse Document Frequency (TF-IDF) vector representations and Bigram counts to evaluate the importance of local transition statistics.

\subsection{Common Training and Evaluation Protocol}
\label{sec:training_setup}

To ensure a rigorous comparison between inductive biases, we enforced a common evaluation protocol across all models. Every experiment utilized the exact same dataset splits ($N_{train}=12,650$, $N_{val}=3,163$) generated with a fixed seed. We report results averaged over five distinct random seeds to rule out initialization luck.

Neural models were trained on a single NVIDIA A100 GPU, while XGBoost and SVM baselines were trained on CPU under identical data and split conditions. To handle class imbalance (Organic $\approx 4\times$ Sybil), we applied inverse-frequency weighting across all architectures: the Cross-Entropy loss was weighted for neural networks, and equivalent sample weights were computed for the XGBoost and SVM objectives.

Table \ref{tab:hyperparameters} details the final hyperparameter configurations for the top-performing variants selected based on validation Matthews Correlation Coefficient (MCC). 

\begin{table}[htbp]
\caption{Hyperparameter Configuration}
\label{tab:hyperparameters}
\centering
\begin{tabular}{lcc}
\toprule
\textbf{Parameter Category} & \textbf{Transformer} & \textbf{XGBoost} \\
\midrule
\textbf{Architecture} & 4 Layers, 4 Heads & 500 Trees \\
\textbf{Hidden Dim / Depth} & $d_{model}=256$, $d_{ff}=2048$ & Max Depth = 8 \\
\textbf{Learning Rate / Eta} & $3e^{-4}$ (AdamW) & $\eta = 0.05$ \\
\textbf{Regularization} & Dropout $p=0.1$, $wd=0.01$ & Subsample = 0.8 \\
\textbf{Batch Size} & 32 (Flat) / 256 (Hier.) & N/A \\
\textbf{Sequence Length} & $L_{max}=256$ tokens & N/A (Aggregated) \\
\bottomrule
\end{tabular}
\end{table}

\section{Experiments}

The experiments evaluate modeling decisions under leakage-aware Ethereum actor classification. We test whether sequential architectures exploit Transaction Grammar better than tree-based models once high-signal shortcuts are removed. We structure the evaluation around four questions: how the dataset is constructed after leakage mitigation, whether actor classes differ in grammatical complexity, how sequential and tabular models compare in classification performance, and whether the resulting models are viable for low-latency deployment.

\subsection{Experimental Setup and Baselines}

We benchmarked five architectures on the same leakage-aware $80/20$ stratified split. The comparison includes:
\begin{itemize}
    \item \textbf{Sequential Models:} a \textit{Flattened Transformer}, a \textit{Hierarchical Transformer}, and a \textit{Hierarchical BiLSTM} to capture temporal dependencies at different input resolutions.
    \item \textbf{Tabular Baseline:} \textit{XGBoost}, using aggregated feature engineering (entropy, counts, timing statistics) to test whether global statistics suffice.
    \item \textbf{Feature-Based Baseline:} \textit{Linear SVM} on TF-IDF and bigram representations to evaluate the importance of local transition statistics.
\end{itemize}

We use MCC as the primary metric, alongside macro-F1 and accuracy, averaged over five runs.  

\subsection{Dataset and Preprocessing}
\label{sec:dataset}

We constructed the dataset by collecting labeled addresses from two sources. Labels for Organic (verified legitimate human users) and Sybil (coordinated bot rings eliminated for reward farming) were sourced from the Hop Protocol's official investigation list \cite{hop_airdrop_github_2026}, while MEV Arbitrage Bots were identified via Dune Analytics. We obtained addresses marked as Arbitrage Bots. For transaction histories that include internal execution traces, we utilized Google BigQuery with a hard cutoff on May 13th, 2022, when Hop Protocol published their snapshot.

We filtered out wallets with fewer than $L_{\min}=10$ transactions to exclude them from the dataset. To mitigate label leakage—defined here as spurious correlations where high-signal proxy contracts trivially reveal the target class rather than underlying behavior—our Blind-Spot protocol removes interactions with entities like OpenSea or Uniswap Router, reducing total volume by 21.8\% (Table~\ref{tab:leakage_removal}). Our dataset now comprises 15,813 sequences (Table~\ref{tab:dataset_stats}), with 80/20 splits for training and validation sets respectively.

\begin{table}[htbp]
\caption{Examples of Top Leakage Sources Removed by the ``Blind-Spot'' Protocol}
\label{tab:leakage_removal}
\centering
\begin{tabularx}{\textwidth}{@{}l l X@{}}
\toprule
\textbf{Contract} & \textbf{Leaked Class} & \textbf{Reason for Removal} \\
\midrule
Uniswap Router & MEV Bot & Circular logic (defines arbitrage). \\
OpenSea Wyvern & Organic & Primary proxy for human NFT (Non-Fungible Token) activity. \\
ENS Controller & Organic & Strong signal of human identity. \\
WETH9 (Wrapped Ether) & Shared / All & Ubiquitous utility (noise reduction). \\
\bottomrule
\end{tabularx}
\end{table}
\FloatBarrier

After removing high-signal contracts, the remaining dataset preserves all three actor categories while reducing shortcut-based separability. Table~\ref{tab:dataset_stats} summarizes the resulting class distribution used for the common train/validation split.

\begin{table}[htbp]
\caption{Dataset Statistics Post-Filtering}
\label{tab:dataset_stats}
\centering
\begin{tabular}{@{}l c c@{}}
\toprule
\textbf{Class} & \textbf{Sequences ($N$)} & \textbf{Split (Train/Val)} \\
\midrule
Organic & 9,790 & 7,832 / 1,958 \\
MEV Bot & 3,710 & 2,968 / 742 \\
Sybil & 2,313 & 1,850 / 463 \\
\midrule
\textbf{Total} & \textbf{15,813} & \textbf{12,650 / 3,163} \\
\bottomrule
\end{tabular}
\end{table}
\FloatBarrier

\subsection{Behavior Grammar Analysis (Q1)}
\label{sec:grammar_analysis}

We investigated whether organic users exhibit higher structural entropy ($H_G$), complexity ($C_{LZ}^G$), and timing entropy ($H_\tau$) than automated bots.

Results show that organic users possess significantly higher $H_G$ (3.50) than MEV (3.12) and Sybil (2.66) actors, with a statistically significant difference (Kruskal-Wallis test $p < 0.01$ and Cliff’s $\delta > 0.3$). Hence, organic users exhibit greater structural complexity (Table~\ref{tab:grammar_metrics}).

\begin{table}[htbp]
\caption{Median Entropy and Normalized LZ Complexity by Class}
\label{tab:grammar_metrics}
\centering
\begin{tabular}{@{}l c c c@{}}
\toprule
\textbf{Metric} & \textbf{Organic} & \textbf{MEV Bot} & \textbf{Sybil} \\
\midrule
Structure Entropy ($H_G$) & \textbf{3.50} & 3.12 & 2.66 \\
Intent Entropy ($H_V$) & \textbf{2.79} & 2.20 & 2.57 \\
Grammar Comp. ($C_{LZ}^G$) & \textbf{3.41} & 3.16 & 2.84 \\
Timing Entropy ($H_\tau$) & \textbf{3.33} & 3.17 & 3.17 \\
\bottomrule
\end{tabular}
\end{table}
\FloatBarrier

Organic users also exhibited the highest timing entropy. However, the difference between MEV and Sybil was negligible (Cliff’s $\delta \approx -0.02$), suggesting that bot timing is shaped by factors beyond actor-specific behavior.

\subsection{Classification Performance (Q2)}
\label{sec:classification_q2}

We addressed the following question: do we really need sophisticated sequence modeling to achieve strong performance, or is a tabular inductive bias sufficient? We tested the hypothesis that XGBoost would outperform sequence models under leakage-aware evaluation, because transaction history is more structured than text-like.

Empirical results strongly support this hypothesis. As shown in Table~\ref{tab:classification_results}, XGBoost outperforms all other models across metrics (MCC: $0.7535$, Macro-F1: $0.8141$), surpassing the Flattened Transformer (MCC: $0.6602$) and the Hierarchical BiLSTM (MCC: $0.6187$) by a substantial margin. Linear SVM also outperforms the BiLSTM. This suggests that local recurrence provides minimal predictive benefit.

 The Flattened Transformer achieved the best recall on bot types MEV and Sybil (MEV $0.90$, Sybil $0.59$), outperforming XGBoost on those two types. But this came at a heavy cost on the ability to generalize well on Organic users: $0.82$ recall compared to XGBoost’s $0.93$.

\begin{table}[htbp]
\caption{Classification Performance (Blind-Spot Dataset)}
\label{tab:classification_results}
\centering
\begin{tabular}{@{}l c c c@{}}
\toprule
\textbf{Model} & \textbf{MCC ($\mu \pm \sigma$)} & \textbf{Macro-F1} & \textbf{Accuracy} \\
\midrule
\textbf{XGBoost (Tabular)} & \textbf{0.7535 $\pm$ 0.0016} & \textbf{0.8141} & \textbf{0.8745} \\
Flattened Transformer & $0.6602 \pm 0.0170$ & $0.7566$ & $0.8151$ \\
Linear SVM & $0.6298 \pm 0.0000$ & $0.7456$ & $0.7946$ \\
Hierarchical BiLSTM & $0.6187 \pm 0.0345$ & $0.7249$ & $0.7625$ \\
Hierarchical Transformer & $0.5733 \pm 0.0227$ & $0.6928$ & $0.7304$ \\
\bottomrule
\end{tabular}
\end{table}
\FloatBarrier

\subsection{Order vs. Timing Ablation (Q3)}
\label{sec:ablation}

To investigate the performance gap between sequence and tabular models, we performed an ablation study by masking specific information in the validation set. We applied three perturbations to the Hierarchical Transformer baseline ($MCC = 0.5733$):

\begin{itemize}
    \item \textbf{Timing Scramble ($\tau$-Ablation):} Timestamp tokens are randomly permuted while maintaining the transaction body sequence.
    \item \textbf{Grammar Shuffle ($\gamma$-Ablation):} Transaction bodies are permuted while maintaining the original temporal rhythm.
    \item \textbf{Full Shuffle:} The entire sequence (Rhythm, Structure, and Intent) is scrambled, destroying all sequential dependencies.
\end{itemize}

\begin{figure}[htbp]
    \centering
    \includegraphics[width=0.50\linewidth]{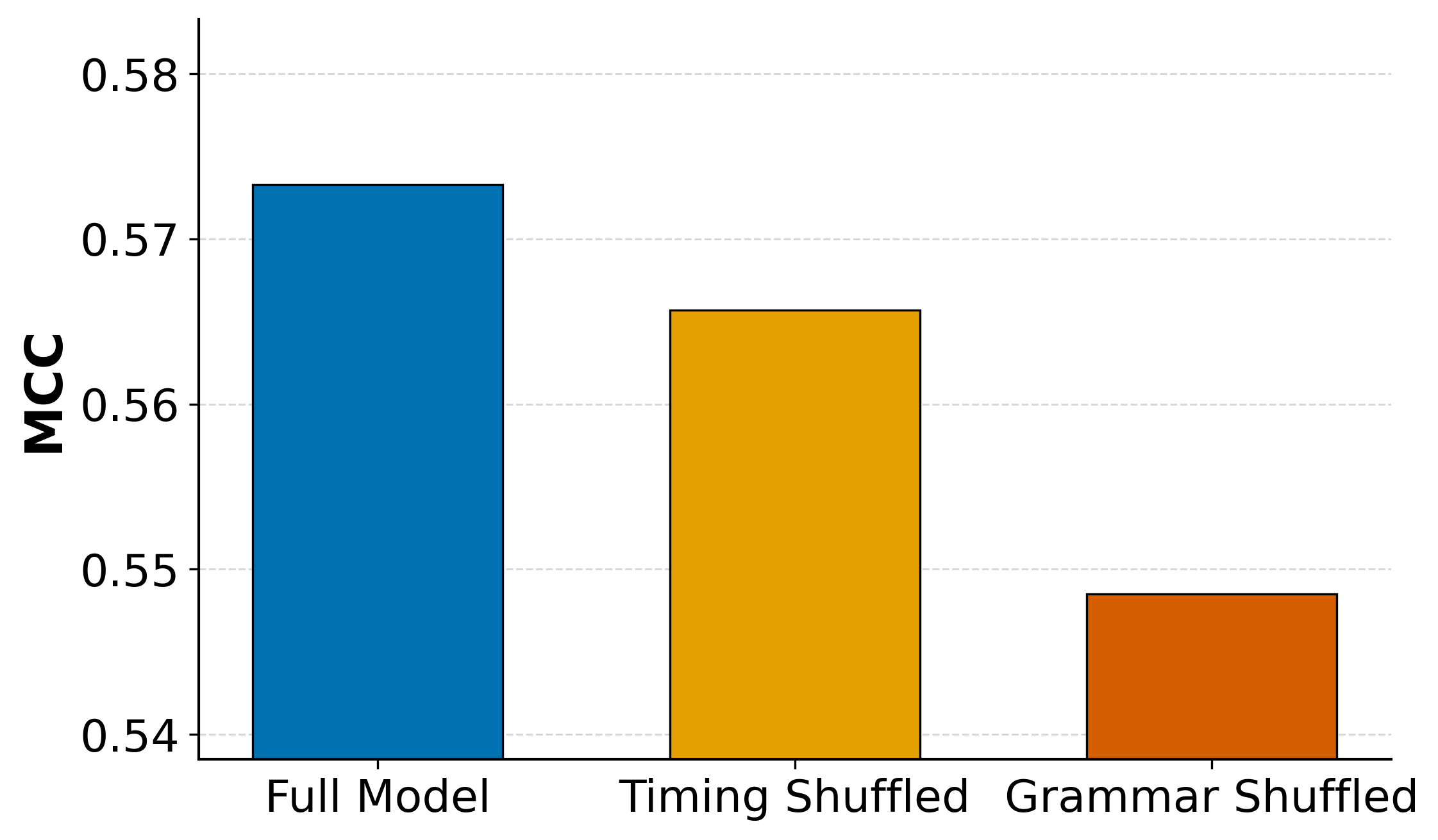}
    \caption{
    Impact of sequence ablations on Transformer performance (MCC). 
    }
    \label{fig:ablation}
\end{figure}

The results (see Fig. \ref{fig:ablation}) show that the model is surprisingly resilient to structural destruction. The Full Shuffle (simulating a Bag-of-Transactions) resulted in an MCC of $0.5485$, a drop of only $\Delta \text{MCC} = -0.0248$. The Grammar Shuffle ($0.5512$) had a significantly larger impact than the Timing Scramble ($0.5657$), indicating that the model derives more signal from the ordering of execution traces than from inter-arrival times.

Critically, the model maintains over 95\% of its predictive power even when the sequence is fully scrambled. This confirms that the Transformer is primarily utilizing local token distributions, effectively acting as a high-dimensional density estimator, rather than capturing long-range temporal dependencies. This explains why XGBoost, which explicitly optimizes for these distribution-based ``Bag-of-Grammar'' features, achieves superior performance with significantly lower complexity.

\subsection{Deployment Viability and Efficiency}
\label{sec:deployment}

\begin{figure}[htbp]
    \centering
    \includegraphics[width=0.95\linewidth, height=0.3\textheight, keepaspectratio]{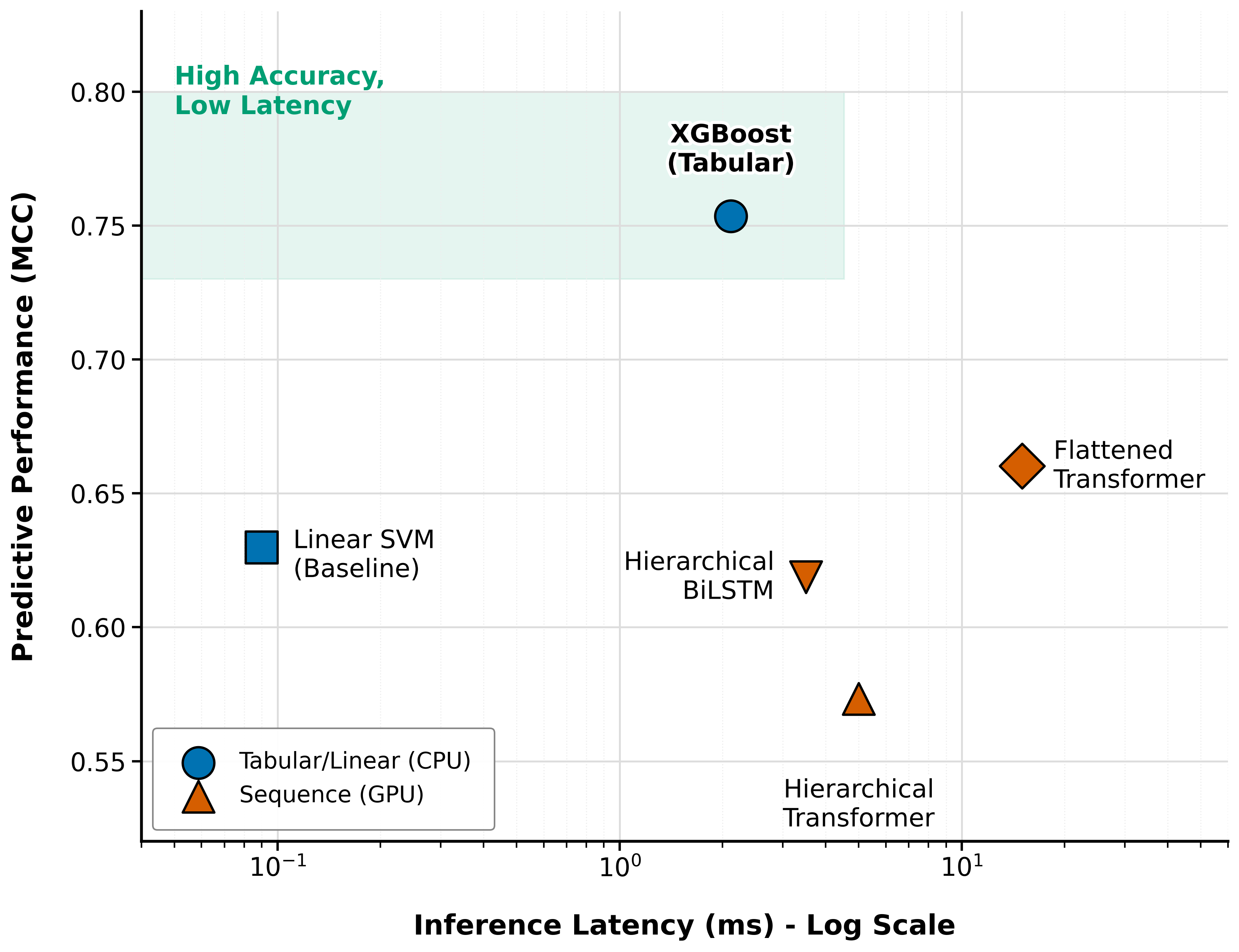}
    \caption{
    Efficiency vs. Accuracy. XGBoost (top left) achieves the most optimal trade-off on the efficiency-accuracy curve. 
    }
    \label{fig:efficiency}
\end{figure}

Fig. \ref{fig:efficiency} shows that the tabular approach is superior in terms of efficiency. This evaluation follows the broader Green AI argument that model quality should be considered together with computational cost and accessibility \cite{schwartz2020green,strubell2019energy}. XGBoost is $100\times$ faster in our experiments, running in microseconds on regular hardware. The Transformer, on the other hand, requires specialized hardware to remain competitive. From the perspective of \textit{Green AI}, the complexity of the Transformer's self-attention at $O(L^2)$ \cite{vaswani2023attentionneed} implies that its energy costs are more than $30\times$ higher than those of the linear complexity of XGBoost. XGBoost dominates the efficiency-accuracy curve because it achieves better predictive performance (Table \ref{tab:classification_results}) while remaining competitive on latency and sustainability.

\FloatBarrier

\section{Discussion and Future Work}

Our results show a mismatch between attention-based inductive biases and blockchain transactions. Because transaction histories lack the compositional semantics of natural language, global sequencing provides less signal than local frequency and execution patterns. Thus, Ethereum behavior is closer to a \textit{Bag-of-Transactions} than to a formal language.

This aligns with tabular-learning evidence that tree-based models often outperform deep models on structured datasets. Structured datasets favor axis-aligned splits, robustness to uninformative features, and feature selection, all naturally handled by gradient-boosted trees. Our results extend this argument to leakage-aware Web3 behavioral analytics.

The \textit{Blind-Spot} approach was essential to this finding. By removing interactions with class-defining contracts, we effectively removed the shortcut features and forced the models to rely on the authentic behavioral signals. With this approach, the tree-based models were able to leverage the global grammatical statistics---entropy, frequency, and dispersion---whereas the sequence models did not capitalize on the long-range structure. 

The results also matter for sustainable and decentralized security. XGBoost improved performance while reducing latency and energy use. Such efficiency enables deployment by validators, light clients, and wallet extensions. Transformer-based models instead favor centralized infrastructure. In Web3, simplicity supports decentralization.

Looking ahead, Graph Neural Networks are promising because Ethereum activity forms a heterogeneous interaction graph. Future work should develop scalable sampling strategies that preserve low latency. Second, Transaction Grammar relies on discrete binning. Continuous embeddings from self-supervised pretraining may test whether richer representations close the gap. Finally, it is essential to be resilient to adversarial camouflage. Clever actors will eventually find ways to adapt their strategies---for example, injecting artificial entropy or random delays---to evade detection. It will be crucial to test models against active evasion attacks to determine whether behavioral classifiers can continue to be effective in the long term.

\section{Conclusion}\label{sec:conclusion}

This paper evaluated Ethereum actor classification under a leakage-aware protocol designed to remove high-signal contract shortcuts before modeling. Using the proposed Transaction Grammar, we showed that organic wallets exhibit higher structural entropy, intent diversity, and grammar complexity than automated actors. However, the classification experiments demonstrated that this behavioral structure is better exploited by aggregate tabular features than by sequence models: XGBoost achieved the strongest MCC, macro-F1, and accuracy, while the ablation study showed that Transformers lose only limited predictive power when transaction order is disrupted. These results support the empirical conclusion that, in this setting, blockchain activity behaves more like a Bag-of-Grammar than a natural-language-like sequence.

The main contribution of the work is therefore a leakage-aware comparison of tree-based and sequential inductive biases for blockchain analytics. The Blind-Spot protocol, Transaction Grammar, order-versus-timing ablation, and efficiency analysis together show that simpler tabular models can be both more accurate and more deployable than heavier neural alternatives, with substantially lower latency and estimated energy use. Future work should extend the same evaluation logic to graph-based representations, self-supervised transaction embeddings, and adversarial camouflage scenarios, while preserving the emphasis on leakage control, computational efficiency, and real-time deployment suitability established in this study.

\bibliographystyle{splncs04}
\bibliography{refs}

\end{document}